\documentclass[10pt,letterpaper]{article}
\usepackage[top=0.85in,left=2.75in,footskip=0.75in]{geometry}

\usepackage{amsmath,amssymb}

\usepackage{changepage}

\usepackage[utf8x]{inputenc}

\usepackage{textcomp,marvosym}

\usepackage{cite}

\usepackage{nameref,hyperref}

\usepackage{microtype}
\DisableLigatures[f]{encoding = *, family = * }

\usepackage[table]{xcolor}

\usepackage{array}

\newcolumntype{+}{!{\vrule width 2pt}}

\newlength\savedwidth

\raggedright
\usepackage[aboveskip=1pt,labelfont=bf,labelsep=period,justification=raggedright,singlelinecheck=off]{caption}

\makeatletter
\renewcommand{\@biblabel}[1]{\quad#1.}
\makeatother

\usepackage{lastpage,fancyhdr,graphicx}
\usepackage{epstopdf}
\fancyheadoffset[L]{2.25in}
\fancyfootoffset[L]{2.25in}
\begin{document}
\vspace*{0.2in}

\begin{flushleft}
{\Large
\textbf\newline{Self-supervision for health insurance claims data: a Covid-19 use case} % Please use "sentence case" for title and headings (capitalize only the first word in a title (or heading), the first word in a subtitle (or subheading), and any proper nouns).
}
\newline
\\
Emilia Apostolova,
Fazle Karim,
Guido Muscioni,
Anubhav Rana,
Jeffrey Clyman

\bigskip
Exponential Technology CoE, Anthem, Chicago, IL, USA
\bigskip

% Insert additional author notes using the symbols described below. Insert symbol callouts after author names as necessary.
% 
% Remove or comment out the author notes below if they aren't used.
%
% Primary Equal Contribution Note

% Additional Equal Contribution Note
% Also use this double-dagger symbol for special authorship notes, such as senior authorship.

% \textcurrency Current Address: Exponential Technology CoE, Anthem, Chicago, IL, USA % change symbol to "\textcurrency a" if more than one current address note
% % \textcurrency b Insert second current address 
% \textcurrency c Insert third current address

% Group/Consortium Author Note
% \textpilcrow Membership list can be found in the Acknowledgments section.

% Use the asterisk to denote corresponding authorship and provide email address in note below.
% \textsuperscript{*} emilia.apostolova@anthem.com

\end{flushleft}
\section*{Abstract}
In this work, we modify and apply self-supervision techniques to the domain of medical health insurance claims. We model patients' healthcare claims history analogous to free-text narratives, and introduce pre-trained `prior knowledge', later utilized for patient outcome predictions on a challenging task: predicting Covid-19 hospitalization, given a patient's pre-Covid-19 insurance claims history. Results suggest that pre-training on insurance claims not only produces better prediction performance, but, more importantly, improves the model's `clinical trustworthiness' and model stability/reliability.

% Please keep the Author Summary between 150 and 200 words
% Use first person. PLOS ONE authors please skip this step. 
% Author Summary not valid for PLOS ONE submissions.   
% \section*{Author summary}
% Lorem ipsum dolor sit amet, consectetur adipiscing elit. Curabitur eget porta erat. Morbi consectetur est vel gravida pretium. Suspendisse ut dui eu ante cursus gravida non sed sem. Nullam sapien tellus, commodo id velit id, eleifend volutpat quam. Phasellus mauris velit, dapibus finibus elementum vel, pulvinar non tellus. Nunc pellentesque pretium diam, quis maximus dolor faucibus id. Nunc convallis sodales ante, ut ullamcorper est egestas vitae. Nam sit amet enim ultrices, ultrices elit pulvinar, volutpat risus.

% Use "Eq" instead of "Equation" for equation citations.
\section*{Introduction}

Self-supervision or pre-training on large unlabeled corpora (word2vec \cite{mikolov2013efficient}, Glove \cite{pennington2014glove}, Elmo \cite{peters2018deep}, Bert \cite{devlin2018bert} and related model, GPT1-3 \cite{radford2018improving,radford2019language,brown2020language}, etc.) has led to continuously improving state-of-the-art results on numerous Natural Language Processing (NLP) tasks. The success of pre-training and self-supervision has recently been expanded to other fields, such as imaging \cite{chen2020generative}, human activity recognition \cite{haresamudram2020masked}, molecular data \cite{rong2020self}, time series clinical data \cite{mcdermott2020comprehensive}, etc.

In this work, we modify and apply NLP self-supervision techniques to the domain of medical health insurance claims (a subset of clinical data). The US health insurance process requires providers (physicians and hospitals) to submit detailed visit claim information for the purposes of health insurance payments. Typically, an insurance claim contains billing codes for various medical diagnoses, procedures and medications, relevant to the billing process. These billing codes are comprised of a subset of the patient's electronic medical record (EMR), and exclude more comprehensive clinical information, such as vital signs and clinical notes. The claims history of a patient can be used for a variety of patient outcome predictions that can help guide and advise patient and provider behaviour for improved health outcomes and healthcare affordability.

We model patients' anonymized health care claims history as a `free-text narrative' and apply self-supervision to introduce prior knowledge, later utilized for patient outcome predictions. The health insurance claim `narrative' consists of a sequence of diagnosis, procedure, and medication codes submitted for billing purposes, together with some basic demographic information, such as age and gender. An example of the information used from a set of anonymized health insurance claims is shown below:

\begin{small}
\textit{
\textbf{Age}: 65; \textbf{Gender}: Female; \textbf{Procedure Code(s)}: G0299 - Direct skilled nursing services of a registered nurse (RN) in the home health or hospice setting, each 15 minutes;
\textbf{Diagnosis Code(s)}: E119 - Type 2 diabetes mellitus without complications; J4590 - Unspecified asthma; Z4881 - Encounter for surgical aftercare following surgery on specified body systems; \textbf{Prescribed Medication Codes}: 0093-5851	Escitalopram; 33342-054	Pioglitazone; 	57664-506	Metoprolol Tartrate; 51248-150	Vesicare.
}
\end{small}

In this study, we focus on utilizing medical health insurance claims pre-training for predicting hospitalization due to a Covid-19 infection, as efforts to reduce mortality due to Covid-19 include early identification and outreach to patients who have the highest risk of developing severe complications from the disease \cite{smith2020distribution}. Predicting post-Covid-19 hospitalization, given patient's pre-Covid-19 insurance claim history is an extremely challenging task due to both the clinical complexity of the disease \cite{wollenstein2020personalized}, as well as the inherently limited and noisy nature of insurance claims (containing only a subset of the patient's EMR, relevant to billing purposes)\footnote{The dataset and clinical outcomes used in this research do not represent the true population of Covid-19 infections due to bias inherent in the dataset gathering process. Results are not indented to provide clinical guidance or clinical analysis of Covid infections.}.

\section*{Related Work}

The work relevant to this study falls into 2 categories: machine learning models focusing on the health insurance claims and self-supervision on clinical data, in particular, data present in insurance claims, such as diagnosis, procedure, and medication codes.

The majority of literature focusing on health insurance claims aims to predict fraud, anomalies, and errors in health insurance claims \cite{kumar2010data, bauder2016predicting,kirlidog2012fraud,shin2012scoring,sowah2019decision,maisog2019using} and typically uses  traditional data mining and machine learning approaches. A few studies focus on predicting medical outcomes from claims. Hung et al. \cite{hung2017comparing} show that a deep neural net and Gradient Boosting Machines (GBM) outperform Support Vector Machines (SVM) and logistic regression on the task of stroke prediction from electronic medical claims. Vekeman et al. \cite{vekeman2019development} use a random forest classifier for identifying patients with Lennox–Gastaut syndrome in health insurance claims. Valdez et al. \cite{valdez2019estimating} analyze the conditions of myalgic encephalomyelitis and chronic fatigue syndrome in insurance claim, and conclude that the symptom information in claims is insufficient to identify diagnosed patients. Nagata et al. \cite{nagata2018prediction} apply GBM and LSTM models to predict risk of type-2 diabetes using claims data.

In terms of self-supervision on clinical data, a number of studies focus on low-dimensional representation learning of medical concepts and medical codes \cite{choi2016learning,choi2016multi,choi2017gram,kartchner2017code2vec,ma2018kame}, utilizing word2vec, Glove, continuous bag-of-words model with time-aware attention, as well as graph-based attention models utilizing medical ontologies. More recently, BEHRT \cite{li2020behrt} applies Bert-like transformer pre-training on Electronic Health Records (EHR) using masked language model that outperforms previous deep EHR representations, such as \cite{nguyen2016mathtt} that combines word2vec embeddings with CNN.  G-BERT is a model that combines Graph Neural Networks and BERT that learn medical representations from MIMIC III. Med-Bert \cite{rasmy2020med} is another BERT-like model that is pre-trained on data from 28 million patients that outperforms BEHRT and G-BERT \cite{shang2019pre}.  We were unable to find studies that focus specifically on self-supervision for medical claims.

\section*{Materials and methods}
\subsection*{Dataset}

In this study, a historical anonymized claims dataset is used to pre-train the model. This dataset contains information of 50 million claims submitted in 2019 and 2020 to a major US health insurance provider. A separate internal dataset that contains 471,971 anonymized Covid-19 positive patients (based on lab result or diagnosis) is used to build a model that would detect patients who are at risk of being hospitalized due to Covid-19 complications. This dataset contains prior 3-years of claim records (diagnosis, procedure, and medication codes) of each Covid-19 positive patient and their respective age and gender. To avoid data leakage, claims up to 7 days prior to a Covid-19 positive diagnosis date are dropped as they may contain information relevant to current Covid-19 infection signs and symptoms. Further, age is discretized into clinically meaningful age ranges \cite{geifman2013redefining}. Covid-19 related hospitalizations were identified based on the primary diagnosis associated with the hospitalization claim. On the other hand, an individual was considered to be not hospitalized, if the individual had non-hospitalized claims subsequent to the COVID-19 positive diagnosis date or the individual did not have any claims 30 days after the COVID-19 positive diagnosis date. The Covid-19 hospitalization rate for the dataset is 15\%. The number is significantly higher than reported in the US \cite{garg2020hospitalization} due to an inherent bias in the dataset which contains patients whose Covid-19 positivity was determined solely by the primary diagnosis of hospitalization. The number is also overestimated by the bias in insurance claims submitted Covid-19 tests (excluding Covid-19 tests without insurance claims and individuals with mild symptoms that were not tested).

\subsection*{Method}

We compare the performance of 4 prediction models on the task of identifying post-Covid hospitalization, given prior 3 years of medical claims history.

As a simple baseline method, we used mappings of diagnosis and procedure codes to the set of known Covid-19 risk factors, e.g. all neoplasm ICD-10 codes (C00-D49) were converted to the risk-factor variable `cancer`. A total of 25 risk factor variables, together with age and gender, were used to build  a logistic regression model on the task. A second baseline method utilizes all available diagnosis, procedure, and drug codes as a bag-of-words representation of the historical claims `narrative' and a Support Vector Machines model \cite{fan2008liblinear}. The baseline methods do not use pre-training and utilize only the dataset of 471,971 Covid-19-diagnosed patients.

The third approach utilizes pre-training on diagnosis, procedure, and medications codes, analogous to word embeddings. Word2vec embeddings\footnote{Continuous bag of words model, with window size 10.} for diagnosis (ICD-10), procedure (Healthcare Common Procedure Coding System), and medication (National Drug Code Directory) codes, each of size 1000, were generated utilizing data from close to 50 million historical claims. The embeddings were then utilized in the Covid-19 positive patients by averaging the embeddings for each type of code respectively (diagnosis, procedure, and medications) for the prior 3 years of the patient's claims history. The 3 types of averaged embeddings were concatenated together with the demographic information (age and gender) and used in a Gradient Boosting Machine (GBM) model \cite{friedman2001greedy} to predict post-Covid-19 hospitalization status.

Lastly, in the forth approach, the dataset of 50 million historical claims was utilized in a transformer-based masked language model: RoBERTa \cite{liu2019roberta}. Before pre-training, data from each of the 50 million claim records were randomly shuffled. Roberta was trained by masking 30\% of the tokens, which include diagnosis/procedure/medications codes, age, and gender. The Roberta model was then fine-tuned on the Covid-19 dataset to predict post-Covid-19 hospitalization status.

Due to the sensitive, clinical nature of the dataset/task and the inherent bias of healthcare claims data, the models' performance needed to be evaluated not only in terms of metrics, such as precision and recall, but also in terms of `clinical trustworthiness' and model stability/reliability. In an attempt to generate explainable model predictions, we applied the LIME feature attribution model \cite{ribeiro2016should} on a random sample of 100 positive and 100 negative predictions for the machine learning models described above. Two clinicians were invited to review the model predictions and determine which model is most clinically `trustworthy' by reviewing explainability results. Unfortunately, due to the size, variability, and both limited and noisy nature of the claims data, the clinicians were not able to utilize the LIME explanations.

As a substitute for human evaluation, we instead measured model stability/reliability by introducing input feature perturbations. For each of our Covid-19 training samples, we substituted each diagnosis/procedure/medication code with the code closest in the corresponding embedding space. Table~\ref{tab:perturb} shows an example of a feature input (medical claim), together with a perturbation automatically generated by substituting each code with the code closest in the pre-trained embedding spaces for diagnoses, procedures and medications. We then measured the differences in prediction probability between the original input and the perturbed input, as well as the differences in the corresponding variable importance scores. The expectations are that such small variations in input should result in minor output / variable significance differences. The perturbations also try to mimic real world coding discrepancies, as medical billing coders have some freedom as to how to code a claim, and the choice of a particular billing code from a set of similar codes is often subjective or circumstantial \cite{lorence2003regional}.

\begin{table}
 \centering

 \begin{tabular}{lll}

 \hline
  \textbf{Input Type} & \textbf{Original Input} & \textbf{Perturbed Input}\\
 \hline
 Medication & \textit{0143988701}  & \textit{0143988775}  \\
 & Amoxicillin 100mL & Amoxicillin 75mL \\
 Procedure & \textit{A7003}  & \textit{A7015}  \\
 & Nebulizer   & Aerosol mask \\
 & administration set & used with nebulizer \\
 Procedure & \textit{1160F}  & \textit{1159F}  \\
 & Review of all   & Medication  \\
 &  medications  &  list documented \\
 Procedure & \textit{99214}  & \textit{99213}  \\
 & Office or oth & Office or oth \\
 & outpatient visit & outpatient visit \\
 Diagnosis & \textit{R062}  & \textit{R06}  \\
 &   Wheezing &   Abnormalities  \\
  &    &   of breathing \\
   Diagnosis & \textit{J189}  & \textit{J18}  \\
   &  Pneumonia,  &  Pneumonia,  \\
  &   unsp organism &   unsp organism \\
 \hline

 \end{tabular}

 \begin{tabular}{lc}
 \hline
 \end{tabular}

 \caption{An example of a feature input (a medical claim), together with a perturbation automatically generated by substituting each code with the code closest in the pre-trained embedding spaces for diagnoses, procedures and medications respectively.}
 \label{tab:perturb}

 \end{table}

The source code for all experiments will be made available at the time of publication. \footnote{Due to compliance regulation we are unable to make publicly available the anonymized claims dataset or derived machine learning models.}.

% Results and Discussion can be combined.
\section*{Results}

Table \ref{tab:result} shows the performance of the four models. 70\% of the 471,971 Covid-19 positive patients were used for training and cross-validation, and the rest 30\% were used for testing. The data used for pre-training consists of claims submitted prior to the first Covid-19 diagnosis in the dataset. As shown, the task proved to be a challenge for all algorithms, with modest precision and recall scores. Results are comparable to results reported in literature utilizing much cleaner, EMR-based datasets \cite{wollenstein2020personalized} on the same task. Clinicians concurred that the task is challenging for human experts, as it is extremely difficult to predict Covid-19 related hospitalizations based solely on the pre-Covid medical history, lacking Covid-related signs, symptoms, and vital signs. The task is further complicated by the noisy and limited nature of medical claims history. Of the two pre-trained models, only the GBM model was able to surpass the SVM and logistic regression baselines.

 \begin{table}

 \centering
 \begin{tabular}{lllll}
 \hline
  & Logit Regres. & SVM & GBM & Roberta\\
 \hline
Precision & 56.0  &  64.5 & 62.8 &  59.3\\
Recall & 54.0 & 50.9 & 55.8 &  49.4 \\
F1 score & 55.3 & 56.9 & 59.1 & 53.8 \\
Accuracy & 86.3 & 88.2 &  88.4 & 89.0 \\
 \hline
 \end{tabular}
 \begin{tabular}{lc}
 \hline
 \end{tabular}
 \caption{Performance of the 4 algorithms: logistic regression on known Covid-19 risk factors, bag-of-words based SVM, pertained embeddings GBM, and the fine-tuned Roberta model.}
 \label{tab:result}
 
 \end{table}

Pre-training, however, seemed to have more significant impact on the `stability', `trustworthiness' of the model and its explainability. Table \ref{tab:perturb_diff} summarizes the prediction probability differences between the original input and the perturbed input, produced by we substituting each diagnosis/procedure/medication code with the code closest in the corresponding embedding space. As individual codes are not used in the logistic regression model, the model was excluded from this evaluation. The differences are summarized in terms of the mean difference between the prediction probability values of the original and perturbed inputs (\textit{Predict Prob Diff Mean}) and in terms of the prediction agreement between the original and perturbed inputs at a probability threshold of 0.5 (\textit{Predict Agreement}). The table also shows the mean squared error computed by comparing the LIME variable importance scores of the original input vs. the perturbed input (\textit{Var Importance MSE}).  Statistics were produced based on 5,000 random samples from the test set. While the baseline bag-of-word SVM approach (without preparing) exhibits the lowest probability output variability, the methods using pre-training exhibit higher prediction agreement on the original vs perturbed inputs, as well as less variability in terms of input variable importance. This could suggest that that the predictions of the pre-trained models are more `stable' in terms of both binary prediction outcome, as well as model explainability.

 \begin{table}

 \centering
 \begin{tabular}{llll}
 \hline
  & SVM & GBM & Roberta\\
 \hline
Predict Prob Diff Mean & 4.93 & 12.90  & 5.31 \\
% Prediction Prob Diff Std Dev & 4.16 & 3.40  & 5.83 \\
% Min & 0.00 & 0.01  & 0.00 \\
% 25\% & 1.76 & 11.81  & 1.54 \\
% Prediction Prob Median & 3.94 & 13.33  & 3.56 \\
% 75\% & 7.07 & 14.82  & 7.04 \\
% Max & 29.37 & 23.11  & 53.06 \\
Predict Agreement & 80.56 & 95.82 & 92.52 \\
Var Importance MSE & 14.81e-4 & 0.85e-4 & 0.98e-4 \\
% Var Importance NDCG & 0.86 & 0.80 & 0.80 \\
 \hline
 \end{tabular}
 \begin{tabular}{lc}
 \hline
 \end{tabular}
 \caption{Differences between the prediction percent probabilities between the prediction of input / perturbed-input pairs for the three algorithms. The row \textit{Predict Agreement} shows the prediction agreement between the original and perturbed inputs at a probability threshold of 0.5. Row \textit{Var Importance MSE} shows the mean squared error of the LIME variable importance of the original vs. the perturbed input.}
 \label{tab:perturb_diff}
 
 \end{table}

Lastly, as a sanity check, we evaluated the 3 model predictions using as input variations of diagnosis and procedure codes for \textbf{all} conditions associated with high risk of Covid-19 hospitalizations \footnote{Centers for Disease Control and Prevention: https://www.cdc.gov/coronavirus/2019-ncov/need-extra-precautions/index.html}, such as cancer, chronic kidney disease, COPD, etc. The logistic regression model was again excluded from this analysis, as the model is explicitly based on known Covid-19 risks. As expected, in all cases the models predicted Covid-19 related hospitalization. However, the SVM baseline model probability averaged at 68\%, while the probability of the pre-trained models was significantly higher, averaging 94\% and 78\% for GBM and Roberta respectively, indicating that the pre-trained models are more confident in predicting such `clear-cut' hospitalization examples. % In terms of variable importance, SVM associated the reported high risk Covid-19 condition of Down syndrome with negative hospitalization, while both pre-trained models associated sickle-cell disease with negative hospitalization. Adults with Down syndrome are at almost five times the risk for COVID-19–related hospitalization  \cite{clift2020covid}, and less for children. While patients with sickle cell disease are almost two times more likely to be hospitalized.

\section*{Conclusion}

This work demonstrated the utility of self-supervision of medical insurance claims data, which can allow Health Insurance Providers to improve ML model performance on a variety of prediction outcome tasks, aiming to improve patient outcomes and health insurance affordability.
Pre-training improved both model prediction performance and model stability on the challenging task of predicting Covid-19 hospitalizations. % Future work involves applying similar pre-training on additional tasks, such as hospital readmission predictions and predicting longitudinal outcomes of a variety of diseases and conditions, including long-term Covid-19 effects.

\bibliographystyle{plos2015}
\bibliography{custom}
% \bibitem{bib1}
% Conant GC, Wolfe KH.
% \newblock {{T}urning a hobby into a job: how duplicated genes find new
%   functions}.
% \newblock Nat Rev Genet. 2008 Dec;9(12):938--950.
%
% \bibitem{bib2}
% Ohno S.
% \newblock Evolution by gene duplication.
% \newblock London: George Alien \& Unwin Ltd. Berlin, Heidelberg and New York:
%   Springer-Verlag.; 1970.
%
% \bibitem{bib3}
% Magwire MM, Bayer F, Webster CL, Cao C, Jiggins FM.
% \newblock {{S}uccessive increases in the resistance of {D}rosophila to viral
%   infection through a transposon insertion followed by a {D}uplication}.
% \newblock PLoS Genet. 2011 Oct;7(10):e1002337.

% \end{thebibliography}

\end{document}